\documentstyle[aclap]{article}

\title{\vspace{-0.5in}Higher--Order Coloured Unification and Natural
  Language Semantics}
\author{Claire Gardent \\
Computational Linguistics \\ Universit\"at des Saarlandes\\
D--Saarbr\"ucken \\
{\tt claire@coli.uni-sb.de}\And
Michael Kohlhase \\
Computer
Science \\ Universit\"at des Saarlandes \\
D--Saarbr\"ucken \\
{\tt kohlhase@cs.uni-sb.de}}

\def\cola{{\sf a}}
\def\colb{{\sf b}}
\def\colc{{\sf c}}
\def\cold{{\sf d}}
\def\cole{{\sf e}}

\def\colpe{{\sf pe}}
\def\colpf{{\sf pf}}
\def\colps{{\sf ps}}

\def\colnpf{{\neg\sf pf}}
\def\colnpe{{\neg\sf pe}}
\def\colnps{{\neg\sf ps}}

\def\colavar{{\sf A}}
\def\colbvar{{\sf B}}

\newcommand{\bc}{\begin{center}}
\newcommand {\ec}{ \end{center}}
\newcommand {\be} {\begin{enumerate}}
\newcommand {\ee} {\end{enumerate}}
\newcommand {\bi} {\begin{itemize}}
\newcommand {\ei} {\end{itemize}}
\newcommand {\ba} {\vspace*{3pt}\noindent$\begin{array}}
\newcommand {\et} {\end{tabular}\vspace*{3pt}}
\newcommand {\bt} {\vspace*{3pt}\noindent\begin{tabular}}
\newcommand {\ea} {\end{array}$\vspace*{3pt}}
\newcommand {\txt}[1]{\mbox{{\it{#1}}}}
\def\wff{\txt{wff}}
\newcommand {\ra}{\rightarrow}
\def\Colours{{\cal C}}
\def\Covars{{\cal C\kern-.2ex V}}
\def\Col{\mbox{$\cal C$}}

\def\fsv#1{\overline{#1}}
\newtheorem{defn}{Definition}[section]

\def\ar{\rightarrow}

\begin{document}
\maketitle
\vspace{-0.5in}
\begin{abstract}
In this paper, we show that Higher--Order Coloured Unification --
a form of unification developed for automated theorem proving --
provides a general theory for modeling the interface between the
interpretation process and other sources of linguistic, non semantic
information. In particular, it provides the
general theory for the Primary Occurrence Restriction which
\cite{DaShPe:eahou91}'s analysis called for.
\end{abstract}

\section{Introduction}
\label{s1}

It is well known that Higher--Order Unification (HOU) can be used to
construct the semantics of Natural Language: \cite{DaShPe:eahou91} --
henceforth, DSP -- show that it allows a treatment of VP--Ellipsis
which successfully captures the interaction of VPE with
quantification and nominal anaphora;
\cite{Pulman:houatiof95,GaKo:fahou96} use HOU to model the
interpretation of focus and its interaction with focus sensitive
operators, adverbial quantifiers and second occurrence expressions;
\cite{GaKoLe:cahou96} shows that HOU yields a simple but precise
treatment of corrections; Finally, \cite{Pinkal:ru95} uses linear HOU
to reconstruct under-specified semantic representations.

However, it is also well known that the HOU approach to NL semantics
systematically over--generates and that some general theory of the
interface between the interpretation process and other sources of
linguistic information is needed in order to avoid this.

In their treatment of VP--ellipsis, DSP introduce an informal
restriction to avoid over--generation: the {\it Primary Occurrence
  Restriction} (POR).  Although this restriction is intuitive and
linguistically well--motivated, it does not provide a general
theoretical framework for extra--semantic constraints.

In this paper, we argue that {\it Higher--Order Coloured Unification}
(HOCU, (cf. sections~\ref{s3},\ref{s6}), a restricted form of HOU
developed independently for theorem proving, provides the needed
general framework. We start out by showing that the HOCU approach
allows for a precise and intuitive modeling of DSP's Primary
Occurrence Restriction (cf. section~\ref{s31}). We then show that the
POR can be extended to capture linguistic restrictions on other
phenomena (focus, second occurrence expressions and adverbial
quantification) provided that the notion of {\it primary occurrence}
is suitably adjusted (cf. section~\ref{s4}). Obviously a treatment of
the interplay of these phenomena and their related notion of primary
occurrence is only feasible given a precise and well--understood
theoretical framework. We illustrate this by an example in
section~\ref{s44}. Finally, we illustrate the generality of the HOCU
framework by using it to encode a completely different constraint,
namely Kratzer's binding principle (cf. section~\ref{s5}).

\section{Higher--Order Unification and NL semantics}
\label{s2}

The basic idea underlying the use of HOU for NL semantics is very
simple: the typed $\lambda$--calculus is used as a semantic
representation language while semantically under--specified elements
(e.g. anaphors and ellipses) are represented by free variables whose
value is determined by solving higher--order equations. For instance,
the discourse (1a) has (1b) as a semantic representation where the
value of $R$ is given by equation (1c) with solutions (1d) and (1e).

\bt{lll}
(1) & a. & {\it Dan likes golf. Peter does too.} \\
& b. & {\it like(dan,golf)}$ \wedge R(peter)$ \\
& c. & {\it like(dan,golf)} $ = R(dan)$ \\
& d. & $R = \lambda x.$ {\it like(x,golf)} \\
& e. & $R = \lambda x.$ {\it like(dan,golf)} 
\et

The process of solving such equations is traditionally called
unification and can be stated as follows: given two terms $M$ and $N$,
find a substitution of terms for free variables that will make $M$ and
$N$ equal. For first order logic, this problem is decidable and the
set of solutions can be represented by a single most general unifier.
For the typed $\lambda$--calculus, the problem is undecidable, but
there is an algorithm which -- given a solvable equation -- will
enumerate a complete set of solutions for this equation
\cite{Huet:auaftlc75}.

Note that in (1), unification yields a linguistically valid solution
(1d) but also an invalid one: (1e). To remedy this shortcoming, DSP
propose an informal restriction, the {\bf Primary Occurrence
  Restriction}:
\begin{quote}
  Given a labeling of occurrences as either primary or secondary, the
  POR excludes of the set of linguistically valid solutions, any
  solution which contains a primary occurrence.
\end{quote}
Here, a {\bf primary occurrence} is an occurrence that is {\it
  directly associated} with a {\it source parallel element}. Neither
the notion of direct association, nor that of parallelism is given a
formal definition; but given an intuitive understanding of these
notions, a {\bf source parallel element} is an element of the source
(i.e. antecedent) clause which has a parallel counterpart in the
target (i.e. elliptic or anaphoric) clause.

To see how this works, consider example (1) again. In this case, {\it
  dan} is taken to be a primary occurrence because it represents a
source parallel element which is neither anaphoric nor controlled
i.e. it is directly associated with a source parallel element.
Given this, equation (1c) becomes (2a) with solutions (2b) and (2c)
(primary occurrences are underlined). Since (2c) contains a primary
occurrence, it is ruled out by the POR and is thus excluded from the
set of linguistically valid solutions.

\bt{lll}
(2) & a.  & $like(\underline{dan},golf) = R(dan)$ \\
& b. & $R = \lambda x. like(x,golf)$ \\
& c. & $R = \lambda x. like(\underline{dan},golf)$ 
\et

Although the intuitions underlying the POR are clear, two main
objections can be raised. First, the restriction is informal and as
such provides no good basis for a mathematical and computational
evaluation. As DSP themselves note, a general theory for the POR is
called for. Second, their method is a generate--and--test method: all
logically valid solutions are generated before those solutions that
violate the POR and are linguistically invalid are eliminated.  While
this is sufficient for a theoretical analysis, for actual computation
it would be preferable never to produce these solutions in the first
place. In what follows, we present a unification framework which
solves both of these problems.

\section{Higher--Order Coloured Unification (HOCU)}
\label{s3}

There is a restricted form of HOU which allows for a natural modeling
of DSP's Primary Occurrence Restriction: Higher--Order Coloured
Unification developed independently for theorem proving
\cite{HuKo:acvotlc95}. This framework uses a variant of the simply
typed $\lambda$-calculus where symbol
occurrences can be annotated with so-called {\em colours} and
substitutions must obey the following constraint:
\begin{quote}
  For any colour constant \colc\ and any \colc --coloured variable
  $V_\colc$, a well--formed coloured substitution must assign to
  $V_\colc$ a \colc --monochrome term i.e., a term whose symbols are
 $\colc$--coloured.
\end{quote}
\subsection{Modeling the Primary Occurrence Restriction}
\label{s31}

Given this coloured framework, the POR is directly modelled as follows:
Primary occurrences are \colpe-coloured whilst free variables are
$\colnpe$-coloured. For the moment we will just consider the colours
$\colpe$ (primary for ellipsis) and $\colnpe$ (secondary for ellipsis)
as distinct basic colours to keep the presentation simple. Only for
the analysis of the interaction of e.g.  ellipsis with focus phenomena
(cf. section \ref{s44}) do we need a more elaborate formalization, which
we will discuss there.

Given the above restriction for well--formed coloured substitutions,
such a colouring ensures that any solution containing a primary
occurrence is ruled out: free variables are $\colnpe$-coloured and
must be assigned a $\colnpe$-monochrome term. Hence no substitution
will ever contain a primary occurrence (i.e. a \colpe-coloured
symbol). For instance, discourse (1a) above is assigned the semantic
representation (3a) and the equation (3b) with unique solution (3c).
In contrast, (3d) is not a possible solution since it assigns to an
$\colnpe$-coloured variable, a term containing a \colpe-coloured
symbol i.e. a term that is not $\colnpe$-monochrome.

\bt{lll}
(3) & a.  & $like(dan_\colpe ,golf) \wedge R_\colnpe (peter)$ \\
& b. & $like(dan_\colpe ,golf) = R_\colnpe (dan_\colpe )$ \\
& c. & $R_\colnpe  = \lambda x. like(x,golf)$ \\
& d. & $R_\colnpe = \lambda x. like(dan_\colpe ,golf)$ 
\et

\subsection{HOCU theory}
\label{s32}

To be more formal, we presuppose a finite set
$\Colours=\{\cola,\colb,\colc,\colpe,\colnpe,\ldots\}$ of {\bf colour
  constants} and a countably infinite supply
$\Covars=\{\colavar,\colbvar,\ldots\}$ of {\bf colour variables}.

As usual in $\lambda$-calculus, the set $\wff$ of {\bf well-formed
  formulae} consists of (coloured\footnote{ Colours are indicated by
  subscripts labeling term occurrences; whenever colours are
  irrelevant, we simply omit them.}) constants
$c_\cola,runs_\colb,runs_\colavar,\ldots$, (possibly uncoloured)
variables $x,x_\cola,y_\colb,\ldots$ (function) {\bf applications} of
the form $MN$ and $\lambda$-abstractions of the form $\lambda x.M$.
Note that only variables without colours can be abstracted over.  We
call a formula $M$ {\bf $\colc$-monochrome}, if all symbols in $M$ are
bound or tagged with $\colc$.
 
We will need the so-called {\bf colour erasure} $|M|$ of $M$, i.e. the
formula obtained from $M$ by erasing all colour annotations in $M$.
We will also use various elementary concepts of the
$\lambda$-calculus, such as {\bf free} and {\bf bound} occurrences of
variables or substitutions without defining them explicitly here. In
particular we assume that free variables are coloured in all formulae
occuring. We will denote the substitution of a term $N$ for all free
occurrences of $x$ in $M$ with $[N/x]M$.

It is crucial for our system that colours annotate symbol occurrences
(i.e.  colours are not sorts!), in particular, it is intended that
different occurrences of symbols carry different colours (e.g.
$f(x_\colb,x_\cola)$) and that symbols that carry different colours
are treated differently.  This observation leads to the notion of
coloured substitutions, that takes the colour information of formulae
into account. In contrast to traditional (uncoloured) substitutions, a
coloured substitution $\sigma$ is a pair
$\langle\sigma^t,\sigma^c\rangle$, where the {\bf term substitution}
$\sigma^t$ maps coloured variables (i.e.  the pair $x_\colc$ of a
variable $x$ and the colour $c$) to formulae of the appropriate type
and the {\bf colour substitution} $\sigma^c$ maps colour variables to
colours. In order to be legal (a {\bf $\Col$-substitution}) such a
mapping $\sigma$ must obey the following constraints:
\begin{itemize}
\item If $\cola$ and $\colb$ are different colours, then
  $|\sigma(x_\cola)|=|\sigma(x_\colb)|$, i.e. the colour erasures have to be equal.
\item If $\colc\in\Colours$ is a colour constant, then $\sigma(x_\colc)$ is
  $\colc$-monochrome.
\end{itemize}
The first condition ensures that the colour erasure of a
$\Col$-substitution is a well-defined classical substitution of the
simply typed $\lambda$-calculus. The second condition formalizes the
fact that free variables with constant colours stand for monochrome
subformulae, whereas colour variables do not constrain the
substitutions. This is exactly the trait, that we will exploit in our
analysis.

Note that $\beta\eta$-reduction in the coloured $\lambda$-calculus is
just the classical notion, since the bound variables do not carry
colour information.  Thus we have all the known theoretical results,
such as the fact that $\beta\eta$-reduction always terminates
producing unique normal forms and that $\beta\eta$-equality can be
tested by reducing to normal form and comparing for syntactic
equality. This gives us a decidable test for {\em validity} of an
equation.

In contrast to this, higher-order unification tests for {\em satisfiability} by finding a
substitution $\sigma$ that makes a given equation $M=N$ valid
($\sigma(M)=_{\beta\eta}\sigma(N)$), even if the original equation is not
($M\ne_{\beta\eta}N$). In the coloured $\lambda$-calculus the space of (semantic) solutions
is further constrained by requiring the solutions to be $\Col$-substitutions. Such a
substitution is called a {\bf $\Col$-unifier} of $M$ and $N$. In particular,
$\Col$-unification will only succeed if comparable formulae have unifiable colours.  For
instance, $intro_\cola(p_\cola,j_\colb,x_\cola)$ unifies with
$intro_\cola(y_\cola,j_\colavar,s_\cola)$ but not with
$intro_\cola(p_\cola,j_\cola,s_\cola)$ because of the colour clash on $j$.

It is well-known, that in first-order logic (and in certain related
forms of feature structures) there is always a most general unifier
for any equation that is solvable at all.  This is not the case for
higher-order (coloured) unification, where variables can range over
functions, instead of only individuals. Fortunately, in our case we
are not interested in general unification, but we can use the fact
that our formulae belong to very restricted syntactic subclasses, for
which much better results are known. In particular, the fact that free
variables only occur on the left hand side of our equations reduces
the problem of finding solutions to higher-order matching, of which
decidability has been proven for the subclass of third-order
formulae~\cite{Dowek:tomid92} and is conjectured for the general case.
This class, (intuitively allowing only nesting functions as arguments
up to depth two) covers all of our examples in this paper.  For a
discussion of other subclasses of formulae, where higher-order
unification is computationally feasible see~\cite{Prehofer:dhoup94}.

Some of the equations in the examples have multiple most general
solutions, and indeed this multiplicity corresponds to the possibility
of multiple different interpretations of the focus constructions. The
role of colours in this is to restrict the logically possible
solutions to those that are linguistically sound.

\section{Linguistic Applications of the POR}
\label{s4}

In section~\ref{s31}, we have seen that HOCU allowed for a simple
theoretical rendering of DSP's Primary Occurrence Restriction. 
But isn't this restriction fairly idiosyncratic? In
this section, we show that the restriction which was originally
proposed by DSP to model VP--ellipsis, is in fact a
very general constraint which far from being idiosyncratic, applies to
many different phenomena. In particular, we show that it is
necessary for an adequate analysis of focus, second occurrence
expressions and adverbial quantification. 

Furthermore, we will see that what counts as a primary occurrence
differs from one phenomenon to the other (for instance, an occurrence
directly associated with focus counts as primary w.r.t focus semantics
but not w.r.t to VP--ellipsis interpretation). To account for these
differences, some machinery is needed which turns DSP's intuitive idea
into a fully--blown theory. Fortunately, the HOCU framework is just
this: different colours can be used for different types of primary
occurrences and likewise for different types of free variables. In
what follows, we show how each phenomenon is dealt with. We then
illustrate by an example how their interaction can be accounted for.

\subsection{Focus}
\label{s41}

Since~\cite{Jackendoff:siigg72}, it is commonly agreed that focus
affects the semantics and pragmatics of utterances. Under this
perspective, {\bf focus} is taken to be the semantic value of a
prosodically prominent element. Furthermore, focus is assumed to
trigger the formation of an additional semantic value (henceforth, the
{\bf Focus Semantic Value} or FSV) which is in essence the set of
propositions obtained by making a substitution in the focus position
(cf. e.g.~\cite{Kratzer:trof91}). For instance, the FSV of
(4a)\footnote{Focus is indicated using upper--case.} is (4b), the set
of formulae of the form {\it l(j,x)} where $x$ is of type $e$,
and the pragmatic effect of focus is to presuppose that the denotation
of this set is under consideration.

\bt{lll}
(4) & a.  & {\it Jon likes SARAH} \\
& b. & $\{l(j,x) \mid x \in\wff_e \}$
\et

In~\cite{GaKo:fahou96}, we show that HOU can successfully be used to
compute the FSV of an utterance. More specifically, given (part of) an
utterance $U$ with semantic representation $Sem$ and foci $F^1 \dots
F^n$, we require that the following equation, the {\bf FSV equation},
be solved:

\ba{l}
 Sem = Gd(F^1) \dots (F^n)
\ea

On the basis of the $Gd$ value, we then define the FSV, written
$\fsv{Gd}$, as follows:

\begin{defn}(Focus Semantic Value) \\
  Let Gd be of type $\alpha =\vec{\beta_{k}}\ra t$ and $n$ be the
  number of foci ($n\leq k$), then the Focus Semantic Value derivable
  from Gd, written $\fsv{Gd}$, is $\{Gd(t^1\dots t^n )\mid
  t^i\in\wff_{\beta_i }\}.$
\end{defn}

This yields a focus semantic value which is in
essence Kratzer's presupposition skeleton.
For instance, given (4a) above, the required equation will be
$l(j,s) = \txt{Gd}(s)$ with two possible values for $Gd$: 
$\lambda x.l(j,x)$ and $\lambda x.l(j,s)$. Given definition (4.1), (4a)
is then assigned two FSVs namely

\bt{lll}
(5) & a. & $\fsv{Gd} = \{l(j,x) \mid x \in\wff_e \} $ \\
& b. & $\fsv{Gd} = \{l(j,s) \mid x \in\wff_e \} $
\et

That is, the HOU treatment of focus over--generates: (5a) is an
appropriate FSV, but not (5b). Clearly though, the POR can be used to
rule out (5b) if we assume that occurrences that are directly
associated with a focus are primary occurrences. To capture the fact
that those primary occurrences are different from DSP's primary
occurrences when dealing with ellipsis, we colour occurrences that are
directly associated with focus (rather than a source parallel element
in the case of ellipsis) $\colpf$.  Consequently, we require that the
variable representing the FSV be $\colnpf$ coloured, that is, its
value may not contain any $\colpf$ term. Under these assumptions, the
equation for (4a) will be (6a) which has for unique solution (6b).

\bt{lll}
(6) & a.  & $\txt{l}(\txt{j,s}_\colpf) = \txt{FSV}_\colnpf (\txt{s}_\colpf$) \\
& b. & $\txt{FSV}_\colnpf  = \lambda x. \txt{l}(\txt{j},x) $
\et

\subsection{Second Occurrence Expressions}
\label{s42}

A second occurrence expression (SOE) is a partial or complete
repetition of the preceding utterance and is characterised by a
de-accenting of the repeating part \cite{Bartels:sot95}. For
instance, (7b) is an SOE whose repeating part {\it only likes Mary} is
deaccented.

\bt{lll}
(7) & a.  & {\it Jon only likes MARY.} \\
& b. & {\it No, PETER only likes Mary.}
\et

In \cite{Gardent:apetdr96,GaKoLe:cahou96} we show that
SOEs are advantageously viewed as involving a deaccented
anaphor whose semantic representation must unify with that of its
antecedent.  Formally, this is captured as follows. Let $SSem$ and
$TSem$ be the semantic representation of the source and target clause
respectively, and $TP^1 \dots TP^n, SP^1 \dots SP^n$ be the target and
source parallel elements\footnote{As in DSP, the
  identification of parallel elements is taken as given.}, then the
interpretation of an SOE must respect the following equations: 

\ba{l} An(SP^1, \dots ,SP^n) = SSem \\ An(TP^1, \dots ,TP^n) = TSem
\ea 

Given this proposal and some further assumptions about the semantics
of {\it only}, the analysis of (7b) involves the following
equations:

\ba{ll}
(8) & An(j) = \forall P [P \in \{\lambda x. like(x,y) \mid y
\in \wff_e \} \\
&\phantom{ An(j) = \forall P [} \wedge P(j) \ra P = \lambda x. like(x,m) ]  \\  
& An(p) =\forall P [P \in FSV \wedge P(p)\\
&\phantom{ An(p) = \forall P [} \ra P = \lambda x. like(x,m) ] 
\ea

Resolution of the first equation then yields two
solutions:

\ba{l}
An = \lambda z \forall P [P \in \{ \lambda x. like(x,y) \mid y \in\wff_e \}\\
\phantom{An = \lambda z \forall P [}\wedge P(z) \ra P = \lambda x. like(x,m) ] \\
An = \lambda z \forall P [P \in \{ \lambda x. like(x,y) \mid y \in
\wff_e \} \\
\phantom{An = \lambda z \forall P [} \wedge P(j) \ra P = \lambda x. like(x,m) ] 
\ea

Since $An$ represents the semantic information shared by target and
source clause, the second solution is clearly incorrect given that it
contains information ($j$) that is specific to the source clause.
Again, the POR will rule out the incorrect solutions, whereby contrary
to the VP--ellipsis case, all occurrences that are directly associated
with parallel elements (i.e. not just source parallel elements) are
taken to be primary occurrences. The distinction is implemented by
colouring all occurrences that are directly associated with parallel
element $\colps$, whereas the corresponding free variable ($An$) is
coloured as $\colnps$. Given these constraints, the first equation in
(8) is reformulated as:

\ba{l}
An_\colnps (j_\colps ) = \forall P [P \in \{\lambda x. like(x,y) \mid y \in \wff_e \}\\
\phantom{An_\colnps (j_\colps ) = \forall P [} \wedge P(j_\colps ) \ra P = \lambda x. like(x,m) ] 
\ea

with the unique well--coloured solution

\ba{l}
An_\colnps  = \lambda z. \forall P [P \in \{ \lambda x. like(x,y) \mid y \in \wff_e \}\\
\phantom{An_\colnps  = \lambda z. \forall P [} \wedge P(z) \ra P = \lambda x. like(x,m) ]  
\ea

\subsection{Adverbial quantification}
\label{s43}

Finally, let us briefly examine some cases of adverbial
quantification. Consider the following example from
\cite{Fintel:amtoaq95}: 

\bt{l}
{\it Tom always takes SUE to Al's mother.} \\
{\it Yes, and he always takes Sue to JO's mother.}
\et

In~\cite{GaKo:fahou96}, we suggest that such cases are SOEs,
and thus can be treated as involving a deaccented anaphor (in this
case, the anaphor {\it he always takes Sue to \_'s mother}). Given
some standard assumptions about the semantics of {\it always}, the
equations constraining the interpretation $An$ of this anaphor are:

\ba{ll}
An(al) = always & \txt{(Tom take x to al's mother)} \\
               &\txt{(Tom take Sue to al's mother)} \\ 
An(jo) = always & FSV \\
& \txt{(Tom take Sue to Jo's mother)}
\ea

Consider the first equation. If $An$ is the semantics shared by target
and source clause, then the only possible value for $An$ is

\ba{ll}
 \lambda z. always &\txt{(Tom take x to z's mother)} \\
                   & \txt{(Tom take Sue to z's mother)}
\ea

where both occurrences of the parallel element $m$ have been abstracted
over. In contrast, the following solutions for $An$ are incorrect.

\ba{ll}
 \lambda z. always & \txt{(Tom take x to al's mother)} \\
                   & \txt{(Tom take Sue to z's mother)} \\ 
 \lambda z. always & \txt{(Tom take x to al's mother)} \\
                   & \txt{(Tom take Sue to al's mother)} \\ 
 \lambda z. always & \txt{(Tom take x to z's mother)} \\
                   & \txt{(Tom take Sue to al's mother)}
\ea

Once again, we see that the POR is a necessary restriction: by
labeling as primary, all occurrences representing a parallel element,
it can be ensured that only the first solution is generated.

\subsection{Interaction of constraints}
\label{s44}

Perhaps the most convincing way of showing the need for a theory of
colours (rather than just an informal constraint) is by looking at the
interaction of constraints between various phenomena. Consider the
following discourse 

\bt{lll}
(9) & a. & {\it Jon likes SARAH} \\
& b. & {\it Peter does too}
\et

Such a discourse presents us with a case of interaction between
ellipsis and focus thereby raising the question of how DSP' POR for
ellipsis should interact with our POR for focus.

As remarked in section~\ref{s31}, we have to interpret the colour
$\colnpe$ as the concept of being not primary for ellipsis, which
includes $\colpf$ (primary for focus).  In order to make this approach
work formally, we have to extend the supply of colours by allowing
boolean combinations of colour constants. The semantics of these
ground colour formula is that of propositional logic, where
$\neg\cold$ is taken to be equivalent to the disjunction of all other
colour constants.

Consequently we have to generalize the second condition on
$\Col$-substitutions
\begin{itemize}
\item For all colour annotations $\cold$ of symbols in
  $\sigma(x_\colc)$ $d\models c$ in propositional logic. 
\end{itemize}
Thus $X_{\neg\cold}$ can be instantiated with any coloured formula
that does not contain the colour $\cold$. The HOCU algorithm is
augmented with suitable rules for boolean constraint satisfaction for
colour equations.

The equations resulting from the interpretation of (9b) are:

\ba{l}
l(j_\colpe ,s_\colpf ) = R_\colnpe (j_\colpe ) \\
R_\colnpe (p) = \txt{FSV}_\colnpf (F)
\ea

where the first equation determines the interpretation of the ellipsis
whereas the second fixes the value of the FSV. Resolution of the
first equation yields the value $\lambda x. l(x, s_\colpf)$ for
$R_\colnpe$. As required, no other solution is possible given the
colour constraints; in particular $\lambda x. l(j_\colpe, s_\colpf)$
is not a valid solution. The value of $R_\colnpe (j_\colpe)$ is now 
$l(p_\colpe, s_\colpf)$ so that the second equation is\footnote{Note
  that this equation falls out of our formal system in that it is
  untyped and thus cannot be solved by the algorithm described in
  section \ref{s6} (as the solutions will show, we have to allow for {\it FSV}
  and $F$ to have different types). However, it seems to be a routine
  exercise to augment HOU algorithms that can cope with type variables
  like \cite{Hustadt:actsfphou91,Dougherty:houuc93} with the colour methods
  from~\cite{HuKo:acvotlc95}.}:

\ba{l}
l(p,s_\colpf) = \txt{FSV}_\colnpf (F)
\ea

Under the indicated colour constraints, three solutions are
possible:

\ba{l}
\txt{FSV}_\colnpf = \lambda x. l(p ,x), F = s_\colpf \\
\txt{FSV}_\colnpf = \lambda O. O(p), F = \lambda
x. l(x,s_\colpf ) \\
\txt{FSV}_\colnpf = \lambda X.X, F = l(p,s_\colpf ) 
\ea

The first solution yields a {\it narrow focus} reading (only {\it
  SARAH} is in focus) whereas the second and the third yield {\it
  wide focus} interpretations corresponding to a VP and an S focus
respectively. That is, not only do colours allow us to correctly
capture the interaction of the two PORs restricting the
interpretation of ellipsis of focus, they also permit a natural
modeling of focus projection (cf.~\cite{Jackendoff:siigg72}).

\section{Another constraint}

\label{s5}

An additional argument in favour of a general theory of colours lies
in the fact that constraints that are distinct from the POR need to be
encoded to prevent HOU analyses from over--generating.
In this section, we present one such constraint (the so-called {\it
  weak--crossover constraint}) and show how it can be implemented
within the HOCU framework. 

In essence, the main function of the POR is to ensure that some
occurrence occuring in an equation appears as a bound variable in the
term assigned by substitution to the free variable occurring in this
equation. However, there are cases where the {\it dual} constraint
must be enforced: a term occurrence appearing in an equation must
appear unchanged in the term assigned by substitution to the free
variable occurring in this equation. The following example illustrates
this.

\cite{Chomsky:corig76} observes that focused NPs pattern with
quantified and wh--NPs with respect to pronominal anaphora: when the
quantified/wh/focused NP precedes and c--commands the pronoun, this
pronoun yields an ambiguity between a co-referential and a
bound--variable reading. This is illustrated in example

\bt{ll}
(10) & {\it We only expected HIM$_i$ to claim} \\
& {\it that he$_i$ was brilliant}
\et

where the presence of the pronoun $he_i$ gives rise to two possible
FSVs\footnote{We abbreviate $exp(x,cl(y,blt(i)))$ to $ex(x,y,i)$
  to increase legibility.}

\ba{l}
FSV =  \{ \lambda x. ex(x,y,i) \mid y \in \wff_e \} \\
FSV =  \{ \lambda x. ex(x,y,y) \mid y \in \wff_e \}
\ea

thus allowing two different readings: the
{\bf corefential} or {\bf strict} reading 

\ba{l}
\forall P[P \in \{ \lambda x. ex(x,y,i) \mid y \in \wff_e\} \\
\phantom{\forall P[}\wedge P(we) \rightarrow P = \lambda x. ex(x,i,i) ]
\ea

and the {\bf bound-variable} or {\bf sloppy} reading. 

\ba{l}
\forall P[P \in \{ \lambda x. ex(x,y,y)) \mid y \in \wff_e\} \\
\phantom{\forall P[}\wedge P(we) \rightarrow P = \lambda x. ex(x,i,i)) ]
\ea

In contrast, if the quantified/wh/focused NP does not precede and
c--command the pronoun, as in

\bt{ll}
(11) & {\it We only expected him$_i$ to claim} \\
& {\it that HE$_i$ was brilliant}
\et

there is no ambiguity and the pronoun can only
give rise to a co-referential interpretation. For instance, given (11)
only one reading arises

\ba{l}
\forall P[P \in \{ \lambda x. ex(x,i,y) \mid y\in \wff_e \} \\
\phantom{\forall P[}\wedge P(we) \rightarrow P = \lambda x. ex(x,i,i) ]
\ea

where the FSV is $\{ \lambda x. ex(x,i,y) \mid y \in \wff_e \}$.

To capture this data, Government and Binding analyses postulate first,
that the antecedent is raised by quantifier raising and second, that
pronouns that are c--commanded and preceded by their antecedent are
represented either as a $\lambda$--bound variable or as a constant
whereas other pronouns can only be represented by a constant (cf. e.g.
\cite{Kratzer:trof91}'s {\it binding principle}).  Using HOCU, we can
model this restriction directly. As before, the focus term is \colpf- and the {\it FSV} variable $\colnpf$-coloured. Furthermore, we
assume that pronouns that are preceded and c--commanded by a
quantified/wh/focused antecedent are variable coloured whereas other
pronouns are $\colnpf$-coloured. Finally, all other terms are taken to
be $\colnpf$-coloured. Given these assumptions, the
representation for (10) is
$ex_\colnpf(we_\colnpf, i_\colpf , i_\colavar)$ 
and the corresponding FSV equation

\bt{l}
$ R_\colnpf (i_\colpf) = \lambda x. ex_\colnpf(x, i_\colpf , i_\colavar)$
\et
 
has two possible solutions

\bt{l}
$R_\colnpf = \lambda y. \lambda x. ex_\colnpf(x, y, i_\colnpf)$  \\
$R_\colnpf = \lambda y. \lambda x. ex_\colnpf(x, y, x)$
\et

In contrast, the representation for (11) is
$ex_\colnpf(we_\colnpf, i_\colnpf , i_\colpf)$  and the equation is

\bt{l}
$R_\colnpf (i_\colpf) = \lambda x. ex_\colnpf(x, i_\colnpf , i_\colpf)$
\et

with only one well--coloured solution

\bt{l}
$R_\colnpf = \lambda y. \lambda x. ex_\colnpf(x, i_\colnpf, y)$
\et

Importantly, given the indicated colour constraints, no other
solutions are admissible. Intuitively, there are
two reasons for this. First, the definition of coloured substitutions
ensures that the term assigned to $R_\colnpf$ is $\colnpf$-monochrome.
In particular, this forces any occurrences of $i_\colpf$ to appear as
a bound variable in the value assigned to $R_\colnpf$ whereas
$i_\colavar$ can appear either as $i_\colnpf$ (a colour variable
unifies with any colour constant) or as a bound variable -- this in
effect models the sloppy/strict ambiguity. Second, a colour constant
only unifies with itself. This in effect rules out the bound variable
reading in (11): if the $i_\colnpf$ occurrence were to become
a bound variable, the value of $R_\colnpf$ would then $\lambda y.
\lambda x. ex_\colnpf(x,y,y)$ . But then by $\beta$--reduction, $R_\colnpf
(i_\colpf )$ would be $\lambda x. ex_\colnpf(x,i_\colpf,i_\colpf)$ which does
not unify with the right hand side of the original equation i.e
$\lambda x. ex_\colnpf(x,i_\colnpf ,i_\colpf) $.

For a more formal account of how the unifiers are calculated see
section~\ref{s61}. 

\section{Calculating Coloured Unifiers}\label{s6}

Since the HOCU is the principal computational device of the analysis
in this paper, we will now try to give an intuition for the
functioning of the algorithm.  For a formal account including all
details and proofs see~\cite{HuKo:acvotlc95}.

Just as in the case of unification for first-order terms, the
algorithm is a process of recursive decomposition and variable
elimination that transform sets of equations into solved forms. Since
$\Col$-substitutions have two parts, a term-- and a colour part, we
need two kinds ($M=^tN$ for term equations and $c=^cd$ for colour
equations). Sets $\cal E$ of equations in solved form (i.e.  where all
equations are of the form $x=M$ such that the variable $x$ does not
occur anywhere else in $M$ or $\cal E$) have a unique most general
$\Col$-unifier $\sigma_{\cal E}$ that also $\Col$-unifies the initial
equation.

There are several rules that decompose the syntactic structure of
formulae, we will only present two of them.  The rule for abstractions
transforms equations of the form $\lambda x.A=^t\lambda y.B$ to
$[c/x]A=^t[c/y]B$, and $\lambda x.A=^tB$ to $[c/x]A=^tBc$ where $c$ is
a new constant, which may not appear in any solution.  The rule for
applications decomposes
$h_\cola(s^1,\ldots,s^n)=^th_\colb(t^1,\ldots,t^n)$ to the set
$\{a=^cb,s^1=^tt^1,\ldots,s^n=^tt^n\}$, provided that $h$ is a
constant.  Furthermore equations are kept in $\beta\eta$-normal form.

The variable elimination process for colour variables is very simple,
it allows to transform a set ${\cal E}\cup\{\colavar=^c\cold\}$ of
equations to $[\cold/\colavar]{\cal E}\cup\{\colavar=^c\cold\}$,
making the equation $\{\colavar=^c\cold\}$ solved in the result. For
the formula case, elimination is not that simple, since we have to
ensure that $|\sigma(x_\colavar)|=|\sigma(x_\colbvar)|$ to obtain a
$\Col$-substitution $\sigma$. Thus we cannot simply transform a set
${\cal E}\cup\{x_\cold=^tM\}$ into $[M/x_\cold]{\cal
  E}\cup\{x_\cold=^tM\}$, since this would (incorrectly) solve the
equations $\{x_\colc=f_\colc,x_\cold=g_\cold\}$. The correct variable
elimination rule transforms ${\cal E}\cup\{x_\cold=^tM\}$ into
$\sigma({\cal
  E})\cup\{x_\cold=^tM,x_{\colc_1}=M^1,\ldots,x_{\colc_n}=^tM^n\}$,
where $\colc_i$ are all colours of the variable $x$ occurring in $M$
and $\cal E$, the $M^i$ are appropriately coloured variants (same
colour erasure) of $M$, and $\sigma$ is the $\Col$-substitution that
eliminates all occurrences of $x$ from $\cal E$.

Due to the presence of function variables, systematic application of
these rules can
terminate with equations of the form
$x_\colc(s^1,\ldots,s^n)=^th_\cold(t^1,\ldots,t^m)$.  Such equations
can neither be further decomposed, since this would
loose unifiers (if $G$ and $F$ are variables, then $Ga=Fb$ as a
solution $\lambda x.c$ for $F$ and $G$, but $\{F=G,a=b\}$ is
unsolvable), nor can the right hand side be substituted for $x$ as in
a variable elimination rule, since the types would clash. Let us
consider the uncoloured equation $x(a)=^ta$ which has the solutions
$(\lambda z.a)$ and $(\lambda z.z)$ for $x$.

The standard solution for finding a complete set of solutions in this
so-called {\bf flex/rigid} situation is to substitute a term for $x$
that will enable decomposition to be applicable afterwards. It turns
out that for finding all $\Col$-unifiers it is sufficient to bind $x$
to terms of the same type as $x$ (otherwise the unifier would be
ill-typed) and compatible colour (otherwise the unifier would not be a
$\Col$-substitution) that either
\begin{itemize}
\item have the same head as the right hand side; the so-called {\bf imitation}
  solution ($\lambda z.a$ in our example) or
\item where the head is a bound variable that enables the head of one of the
  arguments of $x$ to become head; the so-called {\bf projection} binding
  ($\lambda z.z$).
\end{itemize}
In order to get a better understanding of the situation let us reconsider our
example using colours. $x(a_\colc)=a_\cold$. For the imitation solution $(\lambda
z.a_\cold)$ we ``imitate'' the right hand side, so the colour on $\cola$ must be $\cold$.
For the projection solution we instantiate $(\lambda z.z)$ for $x$ and obtain
$(\lambda z.z)a_\colc$, which $\beta$-reduces to $a_\colc$. We see that this ``lifts''
the constant $a_\colc$ from the argument position to the top. Incidentally, the
projection is only a $\Col$-unifier of our coloured example, if $\colc$ and $\cold$ are
identical.

Fortunately, the choice of instantiations can be further restricted to the most general
terms in the categories above.  If $x_\colc$ has type $\overline{\beta_n}\ar\alpha$ and
$h_\cold$ has type $\overline{\gamma_m}\ar\alpha$, then these so-called {\bf general bindings}
have the following form:
\[{\cal G}^h_\cold=\lambda z^{\alpha_1}\ldots z^{\alpha_n}.h_\cold(H_{\cole_1}^1(\overline{z}),\ldots,H_{\cole_m}^m(\overline{z}))\]
where the $H^i$ are new variables of type $\overline{\beta_n}\ar\gamma_i$ and the $\cole_i$ are
either distinct colour variables (if $\colc\in\Covars$) or $\cole_i=\cold=\colc$ (if
$\colc\in\Colours$). If $h$ is one of the bound variables $z^{\alpha_i}$, then
${\cal G}^h_\cold$ is called an {\bf imitation binding}, and else, ($h$ is a constant or a
free variable), a {\bf projection binding}.

The general rule for flex/rigid equations transforms
$\{x_\colc(s^1,\ldots,s^n)=^th_\cold(t^1,\ldots,t^m)\}$ into
$\{x_\colc(s^1,\ldots,s^n)=^th_\cold(t^1,\ldots,t^m), x_\colc=^t{\cal
  G}_\colc^h\}$, which in essence only fixes a particular binding for
the head variable $x_\colc$. It turns out (for details and proofs
see~\cite{HuKo:acvotlc95}) that these general bindings suffice to
solve all flex/rigid situations, possibly at the cost of creating new
flex/rigid situations after elimination of the variable $x_\colc$ and
decomposition of the changed equations (the elimination of $x$ changes
$x_\colc(s^1,\ldots,s^n)$ to ${\cal G}^h_\colc(s^1,\ldots,s^n)$ which
has head $h$).

\subsection{Example}
\label{s61}

To fortify our intuition on calculating higher-order coloured unifiers
let us reconsider examples (10) and (11) with the equations

\ba{l}
R_\colnpf(i_\colpf)=^t \lambda x. ex_\colnpf(x,i_\colpf,i_\colavar) \\
R_\colnpf(i_\colpf)=^t \lambda x. ex_\colnpf(x,i_\colnpf,i_\colpf)
\ea
 
We will develop the derivation of the solutions for the first
equations (10) and point out the differences for the second (11). As a
first step, the first equation is decomposed to

\ba{l}
R_\colnpf(i_\colpf,c)=^tex_\colnpf(c,i_\colpf,i_\colavar)
\ea

where $c$ is a new constant.  Since $R_\colnpf$ is a variable, we are
in a flex/rigid situation and have the possibilities of projection and
imitation. The projection bindings $\lambda xy. x$ and $\lambda xy.y$
for $R_\colnpf$ would lead us to the equations $i_\colpf=^t
ex_\colnpf(c,i_\colpf,i_\colavar)$ and $c=^t
ex_\colnpf(c,i_\colpf,i_\colavar)$, which are obviously unsolvable,
since the head constants $i_\colpf$ (and $c$ resp.)  and $ex_\colnpf$
clash\footnote{For (11) we have the same situation. Here the
  corresponding equation is $i_\colpf=^t
  ex_\colnpf(c,i_\colnpf,i_\colpf)$.}.  So we can only bind
$R_\colnpf$ to the imitation binding $\lambda yx.
ex_\colnpf(H^1_\colnpf(y,x),H^2_\colnpf(y,x),H^3(y,x))$. Now, we can
directly eliminate the variable $R_\colnpf$, since there are no other
variants. The resulting equation

\ba{ll}
    & ex_\colnpf(H^1_\colnpf(i_\colpf,c),H^2_\colnpf(i_\colpf,c),H^3(i_\colpf,c))\\
=^t & ex_\colnpf(c,i_\colpf,i_\colavar)
\ea

can be decomposed to the equations

\ba{ll}
(17) & H^1_\colnpf(i_\colpf,c)=^t c\\
     & H^2_\colnpf(i_\colpf,c)=^t i_\colpf\\
     & H^3_\colnpf(i_\colpf,c)=^t i_\colavar
\ea

Let us first look at the first equation; in this flex/rigid situation,
only the projection binding $\lambda zw.w$ can be applied, since the
imitation binding $\lambda zw.c$ contains the forbidden constant $c$
and the other projection leads to a clash. This solves the equation,
since $(\lambda zw.w)(i_\colpf,c)$ $\beta$-reduces to $c$, giving the
trivial equation $c=^t c$ which can be deleted by the decomposition
rules. 

Similarly, in the second equation, the projection binding
$\lambda zw.z$ for $H^2$ solves the equation, while the second
projection clashes and the imitation binding $\lambda zw.i_\colpf$ is
not $\colnpf$-monochrome.  Thus we are left with the third equation,
where both imitation and projection bindings yield legal solutions:
\begin{itemize}
\item The imitation binding for $H^3_\colnpf$ is $\lambda
  zw.i_\colnpf$, and not $\lambda zw.i_\colavar$, as one is tempted to
  believe, since it has to be $\colnpf$-monochrome. Thus we are left
  with $i_\colnpf=^t i_\colavar$, which can (uniquely) be solved by
  the colour substitution $[\colnpf/\colavar]$.
\item If we bind $H^3_\colnpf$ to $\lambda zw.z$, then we are left
  with $i_\colpf=^t i_\colavar$, which can (uniquely) be solved by the
  colour substitution $[\colpf/\colavar]$.
\end{itemize}
If we collect all instantiations, we arrive at exactly the two possible solutions
for $R_\colnpf$ in the original equations, which we had claimed in section~\ref{s5}:

\ba{l}
R_\colnpf=\lambda yx. ex_\colnpf(x,y,i_\colnpf)\\
R_\colnpf = \lambda yx. ex_\colnpf(x,y,x)
\ea

Obviously both of them solve the equation and furthermore, none is
more general than the other, since $i_\colnpf$ cannot be inserted for
the variable $x$ in the second unifier (which would make it more
general than the first), since $x$ is bound.

In the case of (11) the equations corresponding to (17) are
$H^1_\colnpf(c,i_\colpf)=^tc$, $H^2_\colnpf(c,i_\colpf)=^t i_\colnpf$
and $H^3_\colnpf(i_\colpf)=^t i_\colpf$. Given the discussion above,
it is immediate to see that $H^1_\colnpf$ has to be instantiated with
the projection binding $\lambda zw.w$, $H^2$ with the imitation
binding $\lambda zw.i_\colnpf$, since the projection binding leads to
a colour clash ($i_\colnpf=^ti_\colpf$) and finally $H^3_\colnpf$ has
to be bound to the projection binding $\lambda zw.z$, since the
imitation binding $\lambda zw.i_\colpf$ is not $\colnpf$-monochrome.
Collecting the bindings, we arrive at the unique solution
$R_\colnpf=\lambda yx.ex_\colnpf(x,i_\colnpf,x)$.

\section{Conclusion}
Higher--Order Unification has been shown to be a powerful tool for
constructing the interpretation of NL. In this paper, we have argued
that Higher--Order Coloured Unification allows a
precise specification of the interface between semantic interpretation
and other sources of linguistic information, thus preventing
over--generation. We have substantiated this claim by specifying the
linguistic, extra--semantic constraints regulating the interpretation
of VP--ellipsis, focus, SOEs, adverbial quantification and pronouns
whose antecedent is a focused NP.

Other phenomena for which the HOCU approach seems particularly
promising are phenomena in which the semantic interpretation
process is obviously constrained by the other sources of linguistic
information. In particular, it would be interesting to see whether
coloured unification can appropriately model the complex interaction
of constraints governing the interpretation and acceptability of
gapping on the one hand, and sloppy/strict ambiguity on the other.

Another interesting research direction would be the development and
implementation of a monostratal grammar for anaphors whose
interpretation are determined by coloured unification. Colours are
tags which decorate a semantic representation thereby constraining the
unification process; on the other hand, there are also the reflex of
linguistic, non-semantic (e.g. syntactic or prosodic) information.
A full grammar implementation would make this connection more precise.

\section{Acknowledgements}

The work reported in this paper was funded by the Deutsche
Forschungsgemeinschaft (DFG) in Sonderforschungsbereich SFB--378,
Project C2 (LISA).


\end{document}